\pdfoutput=1

\documentclass[11pt]{article}
\usepackage{subcaption}

\usepackage{emnlp2021}

\usepackage{times}
\usepackage{latexsym}

\usepackage[T1]{fontenc}

\usepackage[utf8]{inputenc}

\usepackage{microtype}

\title{To what extent do human explanations of model behavior align with actual model behavior?}

\author{Grusha Prasad$^\dagger$\thanks{~The work was conducted during an internship at Facebook AI Research.}, Yixin Nie$^\ddagger$, Mohit Bansal$^\ddagger$, Robin Jia$^\star$, Douwe Kiela$^\star$, Adina Williams$^\star$\\
  $^\dagger$ Johns Hopkins University; $^\ddagger$ UNC Chapel Hill; $^\star$ Facebook AI Research\\
  \texttt{grusha.prasad@jhu.edu, adinawilliams@fb.com}
  }

\usepackage{lipsum, babel}

\usepackage{booktabs,graphicx}
\graphicspath{{./images/}} 

\usepackage{adjustbox}
\usepackage{amsfonts}

\usepackage{amsmath}

\newcommand{\phenom}{Importance Alignment}
\newcommand{\reason}{explanation}

\newcommand{\IM}{$\mathbb{I}_m(x,y)$}
\newcommand{\IO}{$\mathbb{I}^O_m(x,y)$}

\newcommand{\hard}{\mathcal{A}^{H}}
\newcommand{\soft}{\mathcal{A}^{S}}
\newcommand{\expert}{\mathcal{A}^{E}}
\newcommand{\threestar}{\textsuperscript{***}}
\newcommand{\twostar}{\textsuperscript{**}}
\newcommand{\onestar}{\textsuperscript{*}}

\begin{document}
\maketitle
\begin{abstract}
Given the increasingly prominent role NLP models (will) play in our lives, it is important for human expectations of model behavior to align with actual model behavior. Using Natural Language Inference (NLI) as a case study, we investigate the extent to which human-generated \reason{s} of models' inference decisions align with how models actually make these decisions. More specifically, we define three alignment metrics that quantify how well natural language \reason{s} align with model sensitivity to input words, as measured by integrated gradients. Then, we evaluate eight different models (the base and large versions of BERT, RoBERTa and ELECTRA, as well as an RNN and bag-of-words model), and find that the BERT-base model has the highest alignment with human-generated explanations, for all alignment metrics. Focusing in on transformers, we find that the base versions tend to have higher alignment with human-generated explanations than their larger counterparts, suggesting that increasing the number of model parameters leads, in some cases, to \textit{worse} alignment with human explanations. Finally, we find that a model's alignment with human explanations is not predicted by the model's accuracy, suggesting that accuracy and alignment are complementary ways to evaluate models.
\end{abstract}

\section{Introduction}

NLP models often make classification decisions in ways humans don't expect them to. For example, Question Answering (QA) models often choose the correct answer for one example, but fail catastrophically on other very similar examples \citep{tulio-etal-2018-semantically, wallace-etal-2019-universal, selvaraju-etal-2020-squinting}, such as answering ``Is the rose red?'' with no, but then ``What color is the rose?'' with ``red'' \citep{tulio-etal-2019-red}. 
VQA models often attend to different portions of images than humans do \citep{das-etal-2016-human}.
NLI models often rely on shallow heuristics, their predictions are inappropriately affected by particular words, and they sometimes perform unexpectedly well from only looking at the hypothesis \citep{gururangan-etal-2018-annotation, poliak-etal-2018-hypothesis, tsuchiya-2018-performance}.
Since people generally do not expect models to base decisions on spurious correlations in the data (cf. \citealt{mccoy-etal-2019-right}), models that make decisions in alignment with human expectations are less likely to make the right decisions for the wrong reasons.

\begin{figure}[t]
    \centering
    \includegraphics[width=\columnwidth]{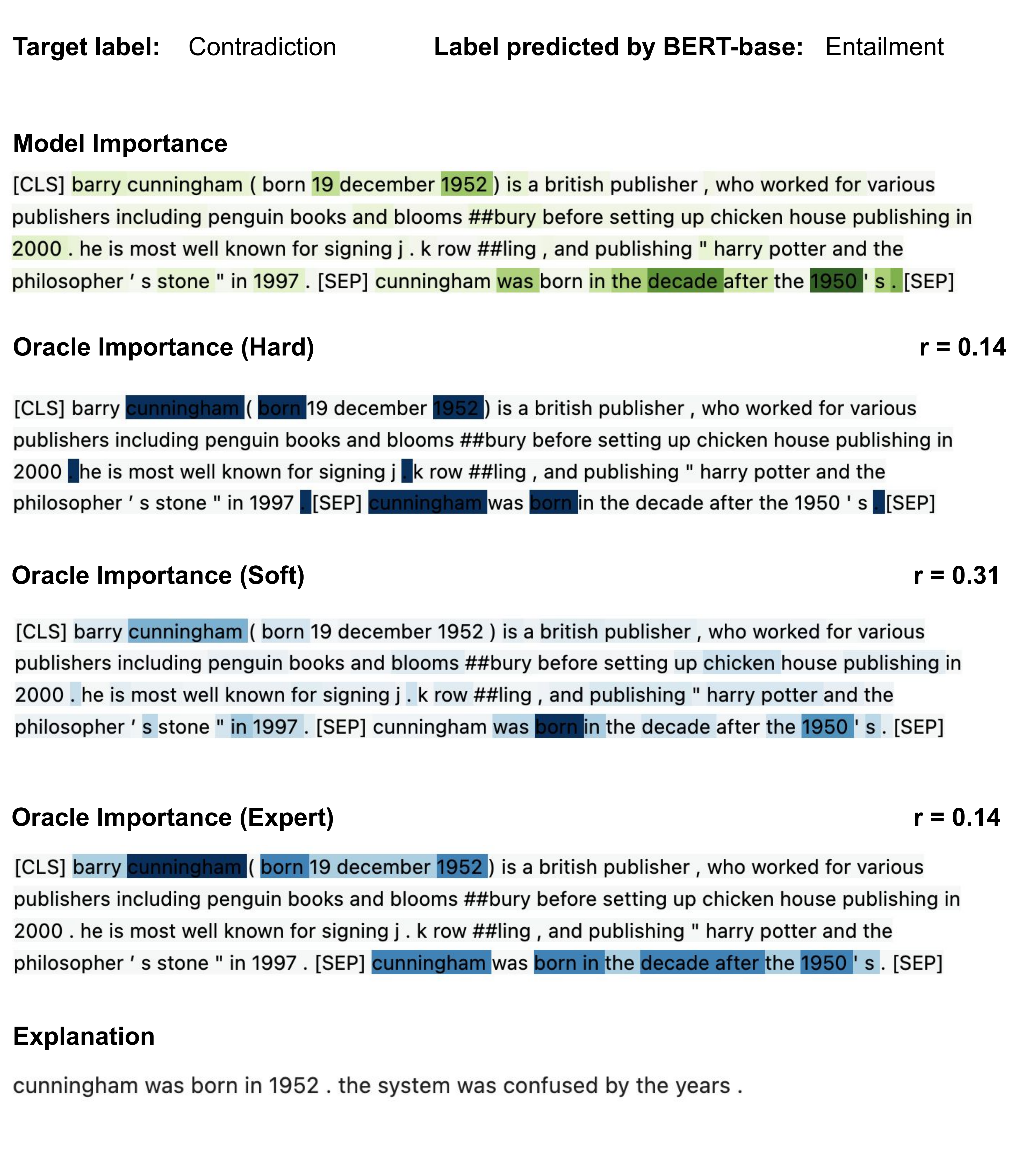}
    \caption{An example illustrating different token-level importance values. ``Model Importance'' is color coded by absolute value integrated gradients attribution for BERT-base. The other three rows show the oracle importance scores estimated by the hard, soft and expert oracles (darker values indicate more important). } 
    \label{fig:example}
\end{figure}

In this paper, we measure how well model decisions are aligned with human expectations about those decisions. Building on work that aims to extract or generate interpretable or faithful descriptions of model behavior \citep{lipton-2018-mythos, rajani-etal-2019-explain, kalouli-etal-2020-xplainli, silva-etal-2020-xte, jacovi-goldberg-2020-aligning, zhao-vydiswaran-2020-lirex}, we use human-generated natural language explanations to determine which portions of the input people expected to be \textit{important} in influencing models' decisions. We then use Integrated Gradients (IG, \citealt{sundararajan-etal-2017-axiomatic}) to determine which portions actually influenced the models' decisions. We term the alignment between them as \textbf{Importance Alignment}. We formulate three different methods of using human-generated natural language explanations to quantify human expectations of model behavior, resulting in three different methods for calculating importance alignment.

As a case study, we applied our method to the Natural Language Inference task \citep{dagan2006, bowman-etal-2015-large, williams-etal-2018-broad} in which models are tasked with classifying pairs of sentences according to whether the first sentence entails, contradicts, or is neutral with respect to the second. Concretely, we measured the extent to which the inference decisions of eight models (six state-of-the-art transformers, an LSTM model and a bag-of-words model) aligned with human-generated explanations from the Adversarial NLI dataset  (ANLI,  \citealt{nie-etal-2020-adversarial}). 

In all three methods for calculating Importance Alignment, BERT-base had the highest importance alignment score. We also found that the smaller, `base' versions of transformers tended to have higher importance alignment scores than the corresponding large versions. However, being smaller doesn't always result in higher importance alignment, since both small non-transformer models had lower importance alignment scores than `base' transformers. Finally, we demonstrate that more accurate models (both for classic test accuracy and on the diagnostic dataset HANS; \citealt{mccoy-etal-2019-right}) do not necessarily have higher importance alignment, suggesting that accuracy and alignment with human expectations are orthogonal dimensions along which models should be evaluated.

\section{Related Work}

The term ``alignment'' has been used in several different contexts in AI: alignment of model behaviour with normative notions of human ethics (``value alignment''; \citealt{russell2015research, peng-etal-2020-reducing}), alignment between tokens from source to target in machine translation, alignment between images and text in image-caption alignment models, etc. In this paper, we propose a new type of alignment, \phenom: we want models to not only generate accurate outputs, but also to generate these accurate outputs for reasons that align with human expectations.

High importance alignment can be valuable because prior work has demonstrated that when people form correct mental models of AI decision boundaries, they make better AI-assisted decisions \citep{bansal-etal-2019-beyond,bansal2019updates}. For example, when annotators are provided with additional information about model decisions, such as model accuracy (cf. \citealt{yin-etal-2019-understanding}) or model-generated explanations \citep{bansal-etal-2020-whole}, it increases their level of trust and in some settings, can actually improve AI-assisted decision making \citep{zhang2020effect}. Therefore, optimizing models to have high importance alignment is a worthy goal, even if it can initially result in models with lower accuracy. In the words of \newcite{bansal-etal-2020-optimizing}, ``predictable performance is worth a slight sacrifice in AI accuracy,'' especially on tasks with potentially serious social implications.

\section{Measuring \phenom}

We assume that for some input example $x = \{x_1, x_2 ... x_n\}$ with $n$ tokens and a gold label $y$, there exists an annotator-generated explanation of why the gold label is correct and/or why a model might output an incorrect prediction. We convert this natural language explanation into an \textbf{oracle importance score} (\IO) which quantifies the extent to which annotators expect each token in $x$ to push the model's prediction towards or away from the gold label.\footnote{We refer to this as an ``oracle'', because we consider importance scores derived from human-generated explanations to be the ground truth.} Then, to compute \phenom\ we correlate the oracle importance score for $x$ with the \textbf{model importance} (\IM), which quantifies the extent to which each token in $x$ actually pushes the model's prediction towards or away from the gold label. A greater correlation between model and oracle importance scores indicates a greater alignment between how annotators expect models to make decisions and how these models actually make decisions. In the remainder of this section we describe our methods to calculate oracle and model importance scores as well as the importance alignment metric from correlating these two scores.

\subsection{Computing model importance scores}
\label{subsec:computingModImp}

We compute model importance scores using \textit{Integrated Gradients} (IG; \citealt{sundararajan-etal-2017-axiomatic}). Concretely, we define model importance ($\mathbb{I}$) for some model $m$ and some example $x$ (e.g., the concatenation of the premise and hypothesis for NLI) with the gold label $y$, as follows,
\begin{align}
\label{eq:model_importance}
\mathbb{I}_m(x,y) = |IG_m(x, y)|
\end{align}
where $IG_m$ returns a vector of IG attribution scores with respect to the gold label for each token in $x$ and $|\cdot|$ denotes component-wise absolute value.

For some token in the input $x_i$, a positive IG attribution score indicates that $x_i$ pushed the model's prediction towards the gold label, whereas a negative IG attribution score indicates that $x_i$ pushed the model's prediction away from the gold label. In a model with high importance alignment, we would expect positive attribution scores to be correlated with annotator expectations about why the gold label is correct, and negative attribution scores to be correlated with annotator expectations about why a model might output an incorrect prediction. In this paper, we are considering explanations which capture both of these annotator expectations without differentiating between them. Therefore, we use the absolute value of the IG attribution score. 

\paragraph{Why Integrated Gradients?}  We use Integrated Gradients because they are axiomatically both interpretable and faithful \cite{sundararajan-etal-2017-axiomatic}, unlike attention based methods which have been argued are not faithful explanations of models' decision making processes (\citealt{jain-wallace-2019-attention}, but see \citealt{wiegreffe-pinter-2019-attention} for a counterpoint). Other perturbation methods which are more faithful than attention such as LIME \citep{ribeiro-etal-2016-lime}, SHAP \citep{lundberg-lee-2016-shap}, and their variants can be used, although these methods have been argued to be unreliable \citep{camburu-etal-2019-can, slack-etal2020-fooling, camburu-etal-2021-struggles}. No current consensus exists on which methods should be employed \citep{hase-bansal-2020-evaluating, Ghorbani_Abid_Zou_2019}. Although we use IG, crucially, our method is not dependent on it; IG can be replaced with any method that can faithfully assign an importance score to each token in the input.

\subsection{Computing oracle importance scores}

We describe three methods of converting natural language explanations into oracle importance scores: hard oracle, soft oracle and expert oracle. 

\paragraph{Hard oracle importance.} 

Hard oracle importance is a binary measure of token overlap between the input and the explanation. This measure captures the intuition that if annotators thought that specific tokens in the input are important for pushing the model towards or away from the gold label, then they will use those words in their explanation. Formally, for some input $x$ with gold label $y$ and explanation $e$, we define hard oracle importance as follows, where the overlap function yields a binary vector in which the $i$-th component is valued $1$ if $x_i \in e$ and $x_i$ is not a stop word;\footnote{We used the list of stop words from NLTK \citep{nltk}} the $i$-th component of the vector is valued $0$ otherwise.
\begin{align}
\mathbb{I}_m^{H}(x,y,e) = overlap(x_m,e)
\end{align}
This measure is model specific only to the extent that the models differ in how they tokenize the input $x$; we assume that the annotator generated explanations themselves do not describe expectations about specific models. 

\paragraph{Soft oracle importance.}
The hard oracle is very simple and does not capture synonyms, entities referred to with pronouns, paraphrases, etc. To overcome these shortcomings, we also define soft oracle importance where we compute importance scores from IG of explanation-informed models. The input for these explanation-informed models is the original input $x$ concatenated with some explanation $e$. The task of the model is to predict not only the gold label for the original task $y$, but also a binary output indicating whether $e$ was an explanation about the current example or about some other example. This task requires the model to perform not only the target task (e.g., NLI) but also requires the model to establish a relationship between the provided explanation and the input, thereby incorporating information from the natural language explanation. 

Formally, for some input $x$ with gold label $y$ and explanation $e$, we define soft oracle importance as,
\begin{align}
\mathbb{I}_m^{S}(x,y,e) = |IG_{m^{'}}(x, e, y^{'})|
\end{align} 

where $m^{'}$ refers to the explanation-informed model and $y^{'}$ refers to the target output of  $m^{'}$ which incorporates both $y$ and a binary output indicating whether $e$ is a matched explanation (e.g., for NLI, $y^{'}$ would have six possible values).  This measure is also model specific both because of tokenization and because the explanation-informed model has the same model architecture as the target model. 

\paragraph{Expert oracle importance.}
The hard and soft oracles are automatic ways of computing oracle importance scores. To validate these automatic measures, we also computed oracle importance scores from experts (three of the authors on this paper).  Given the input, gold label, and annotator generated explanation for a given example, the expert annotators ($N=3$) indicated which tokens of the input they believed that the original annotator (i.e., the one who generated the explanation) thought were important for the model's prediction. Since generating expert annotations was very time consuming, we computed this measure only for a random subset of $60$ examples.  

Formally, for some input $x$ with gold label $y$ and explanation $e$, we define expert oracle importance for any given token as the proportion of expert annotators who indicated that the token was important according to the annotator who generated the explanation. This is expressed as, 
\begin{align}
\mathbb{I}_m^{E}(x,y,e) = \dfrac{1}{N}\sum_{k}^{N} expert_{k}(x_m,e,y)
\end{align}
where $expert_k$ returns a binary vector in which the $i$-th element is valued $1$ if annotator $k$ indicated that the $i$-th token was important, and $0$ otherwise.

Like with hard oracle importance, this measure is model specific only to the extent that the models differ in how they tokenize the input $x$. 

We could not compute oracle importance scores from the original annotators and had to rely on expert annotators because this information was absent from the dataset of natural language explanations we used. Additionally, we argue below that collecting high-quality oracle importance annotations from na\"ive annotators can be very tricky.

\paragraph{Why start from human-generated natural language explanations?} We convert natural language explanations to oracle importance scores instead of collecting oracle importance scores directly from na\"ive annotators for two reasons. First, there already exist data sets of natural language explanations, where annotators were required to reason about models' decision making in an adversarial setting \cite{nie-etal-2020-adversarial}, and more such data sets are being generated \cite{kiela-etal-2021-dynabench}. Second, we contend that for most non-expert annotators, asking them to provide verbal descriptions is easier and more natural than asking them to answer a question like, ``For which words do you think the model's prediction would change the most if that word was blanked out?''. In fact, to pursue this angle assiduously, one would ideally recruit annotators who know what IG is and ask them to predict IG scores---after all, since we are using IG scores to quantify how the models make decisions, the best way to quantify ``how humans think models make decisions'' would be to measure what humans think the IG scores will be. Unfortunately, such a task would be challenging for most non-expert annotators, making it infeasible to collect high quality annotations. 

\subsection{Importance Alignment Metric}
\label{subsec:IAmetric}

For each example with input $x$, gold label $y$ and explanation $e$, we compute importance alignment for some model $m$ as the mean Fisher-transformed product-moment (i.e., Pearson's) correlation ($r$) between the model importance and oracle importance scores for that example as below, where the oracle $O$ is either the hard ($H$), soft ($S$) or expert ($E$) oracles:
\begin{align}\label{eq:cor}
C_m(x, y) = \operatorname{arctanh}(r(\mathbb{I}_m(x,y), \mathbb{I}_m^{O}(x,y,e)))
\end{align}

\noindent We Fisher-transform the correlation coefficient $r$ to ensure that  $C_m(x, y)$ is unbiased and does not violate normality assumptions required for the statistical analyses we use \cite{fisher-1921-probable}.\footnote{Fisher transformed correlation coefficients are approximately normally distributed when the correlation coefficients are calculated from sample pairs drawn from bivariate normal distributions. Although $\mathbb{I}_m(x,y)$ and $\mathbb{I}_m^{O}(x,y,e))$ are not normally distributed, visual examination of the resulting fisher transformed correlations revealed that $C_m(x, y)$ were in fact approximately normal.}

For each example, we also compute a random baseline for $C_{m}(x, y)$ where the oracle importances are calculated by pairing the input with an explanation ($e_R$) which was written for a different input example and was chosen at random:
\begin{align}\label{eq:cor_rand}
C_{m_R}(x, y) = \operatorname{arctanh}(r(\mathbb{I}_m(x,y), \mathbb{I}_m^{O}(x,y,e_R)))
\end{align}
We compute this measure to control for spurious patterns that can drive the correlation between model and oracle importances: we can find a non-zero correlation between the importance scores if a certain model $m$, such as BERT, (explanation informed or not) always assigned high IG attribution values to tokens at specific indices; we can also find a non-zero correlation if certain tokens (e.g., ``the'') received high attribution irrespective of the context, and these tokens occurred frequently in the input and explanation. 

To measure the extent to which model and oracle importances are correlated with each other over and above spurious correlations, we measure the mean difference ($\Delta \mathcal{A}$) between $C_{m}(x, y)$ and $C_{m_R}(x, y)$ for all examples in some dataset $D$ and back-transform it to the correlation scale:

{\small
\vspace{-1em}
\begin{align}
    \Delta \mathcal{A} = \operatorname{tanh} \left(\frac{1}{\mid D \mid} \sum_{(x, y) \in D} C_{m}(x, y) - C_{m_R}(x, y)\right)
\end{align}
}%
To measure whether $\Delta \mathcal{A}$ is significantly greater than $0$, we use paired t-test between $C_{m}(x, y)$ and $C_{m_R}(x, y)$ for all examples in $D$. We compute a different measure of $\Delta \mathcal{A}$ for each oracle importance score.

\section{Experimental details}

\subsection{Models}

\paragraph{Target models.}
We measured the importance alignment for six pretrained Transformer language models: BERT base and large \citep{devlin-etal-2019-bert};
RoBERTa base and large \citep{liu-et-al-2019-roberta}; and ELECTRA base and large \citep{clark-etal-2020-electra}. We fine-tuned these models on the combination of the following NLI datasets: SNLI \cite{bowman-etal-2015-large}, MultiNLI \cite{williams-etal-2018-broad}, NLI-recast FEVER \citep{thorne-etal-2019-fever2} and ANLI rounds 1--3 \cite{nie-etal-2020-adversarial}. We used the hyperparameters in the ANLI codebase for fine-tuning.\footnote{\url{https://github.com/facebookresearch/anli/blob/master/script/example_scripts/}} 

We used the same datasets to also train two non-transformer models: a bag-of-words (BOW) model and a RNN based model we call \textbf{BInferSent} for `BERT-InferSent'. The BOW model has a single max pooling layer on top of BERT token embeddings. The BInferSent model combines the InferSent architecture of \newcite{conneau-etal-2017-supervised} with Short-Stacked Sentence Encoders of \newcite{nie-bansal-2017-shortcut}, using 3 layers of BiLSTMs ~\citep{hochreiter-schmidhuber-1997-long} with residual connections on top of BERT token embeddings.

\begin{table}[t]
\centering
\begin{adjustbox}{max width=\linewidth}
\small
    \begin{tabular}{lcc}
    \toprule
          &  Target  &  Explanation informed  \\ 
          & models & models \\
   \midrule
  BERT-Base         & 48.02     & 44.85 \\
  RoBERTa-Base      & 50.47     & 60.93 \\
  ELECTRA-Base      & 52.33     & 51.93 \\
  BERT-Large        & 49.24     & 49.01 \\
  RoBERTa-Large     & 55.37     & 74.21 \\ 
  ELECTRA-Large     & 58.06     & 74.93 \\
  BInferSent        & 40.00     & 20.97 \\
  BOW               & 35.82     & 18.70 \\

     \bottomrule
    \end{tabular}
    \end{adjustbox}
        \caption{Accuracy on the development partition of the ANLI dataset for target models (finetuned on MNLI + SNLI + ANLI + re-cast FEVER) and models used as the soft oracle (finetuned on 6-way NLI and reason classification on a subset of ANLI). }\label{tab:devacc}
\end{table}

\paragraph{Explanation informed models.}
To compute soft oracle importance, we trained explanation informed models for each target model architecture on a six-way classification task. We trained three models per architecture using different random seeds. In this task, the input to the model was a context-hypothesis pair concatenated with an annotator-generated explanation. The output of the model was a joint label indicating whether the context entails, contradicts or is neutral with respect to the hypothesis, and whether the explanation matches the context-hypothesis pair.

We generated the training ($n = 19043$) and development ($n = 2116$) datasets for these classifiers by subsetting the portion of the ANLI training set for which an explanation was provided --- i.e., the examples in the training set in which the ANLI annotators had successfully fooled the model. We presented each of the $19043$ examples twice when training the explanation informed models: once with a matched explanation (i.e., the original one written by the annotator for that explanation) and once with a randomly selected explanation (see \S\ref{subsec:IAmetric} above). We trained all models for two epochs. 

The accuracy of these explanation informed models either matched or surpassed that of the target model despite being trained on only a small subset of the original data (see \autoref{tab:devacc}). This suggests that these models did learn to incorporate information about explanations, and that the information present in the explanations was useful for NLI.

\begin{table}[t]
\centering
\begin{adjustbox}{max width=\linewidth}
    \begin{tabular}{llll}
    \toprule
    \bf Model & \multicolumn{2}{c}{\textbf{Importance Alignment}} & \multicolumn{1}{c}{\textbf{Acc.}} \\
           &  $\Delta\hard$ &  $\Delta\soft$ & ANLI \\ 
   \midrule
  BERT-Base    & 0.21\textsuperscript{***}  & 0.11\textsuperscript{***}& 48.02 \\ 
  RoBERTa-Base & 0.11\textsuperscript{***}  & 0.02\textsuperscript{*} & 50.47 \\ 
  ELECTRA-Base  & 0.17\textsuperscript{***}  & 0.06\textsuperscript{***} & 52.33\\ 
  BERT-Large   & 0.18\textsuperscript{***}  & -0.02 & 49.24 \\ 
  RoBERTa-Large  & 0.04\textsuperscript{*}   & 0.01                    & 55.37 \\ 
  ELECTRA-Large   & 0.07  & 0.01                     & 58.06 \\  
     \midrule
  All Base Trans.   & 0.17  & 0.07 & 50.27\\ 
  All Large Trans.  & 0.11  & -0.003 & 54.56\\ \midrule 
   BInferSent  & 0.12\textsuperscript{***}  & $<$0.01 & 40.00 \\ 
   BOW  & 0.01\textsuperscript{***}  & 0.01\textsuperscript{***} & 35.82 \\ 
     \bottomrule
    \end{tabular}
    \end{adjustbox}
        \caption{Importance Alignment between model importance scores and oracle importance scores (both $\hard$ and $\soft$ metrics) across 5 random seeds on the ANLI dataset. \textbf{$\Delta\mathcal{A}$ was computed over the examples that the models got wrong}. Average model accuracy across seeds and different rounds of ANLI is also provided. `*'s indicate whether $\Delta\mathcal{A}$ is significantly greater than $0$. `***' indicates $p < 0.001$, `**' indicates $p < 0.01 $ and `*' indicates $p < 0.05$. }\label{tab:results}
\end{table}

\subsection{Evaluation Datasets}

\paragraph{ANLI.}
We measured importance alignment on the development set of the ANLI dataset. In this dataset, annotators were given a context and a label and were asked to write a hypothesis that fooled a target model; if the model was fooled, annotators explained in natural language why the provided label was correct and why they thought model was fooled into generating an incorrect prediction. While other datasets with natural language explanations for NLI exist, (e.g., e-SNLI; \citealt{camburu-etal-2018-ensli}), the explanations in these datasets only address why the gold label is correct, and not why the annotators thought the model generated an incorrect prediction. Additionally, the adversarial setting in which ANLI was collected encourages the annotators to reason about models' decision making. These two factors make ANLI more suitable for our purposes than other explanation datasets.

Since annotators provided explanations only when the model generated an incorrect prediction, to  allow for an apples-to-apples comparison, we only compute importance alignment for examples in which our target models generate an incorrect prediction. As a consequence, the specific examples used to measure importance alignment differs across models: different models fail on different examples. To ensure that our results are not driven by the differences between examples, we repeated our analyses on a subset of examples that all models failed on, and we found that our results were nearly identical (compare \autoref{tab:results} and \autoref{tab:results_subset}).

We collected expert annotations (see \S\ref{subsec:computingModImp}) on $60$ randomly sampled examples from the set of examples that all the models failed on, $20$ from each round of ANLI. We computed the expert importance alignment score ($\mathcal{A}^E$) for this subset of examples and repeated our analyses.

\begin{table}[t]
\centering
\begin{adjustbox}{max width=0.93\linewidth}
    \begin{tabular}{llll}
    \toprule
    \bf Model & \multicolumn{3}{c}{\textbf{Importance Alignment}} \\
           &  $\Delta\hard$ &  $\Delta\soft$ & $\Delta\expert$  \\ 
   \midrule
  BERT-Base    & 0.21\textsuperscript{***}  & 0.14\textsuperscript{***} & 0.39\textsuperscript{***}  \\ 
  RoBERTa-Base & 0.10\textsuperscript{***}  & 0.04\textsuperscript{***} & 0.31\textsuperscript{***}  \\ 
  ELECTRA-Base  & 0.15\textsuperscript{***}  & 0.09\textsuperscript{***} & 0.33\textsuperscript{***} \\ 
  BERT-Large   & 0.15\textsuperscript{***}  & -0.03 &  0.27\textsuperscript{***} \\ 
  RoBERTa-Large  & 0.02  & -0.01 & 0.21\textsuperscript{***} \\ 
  ELECTRA-Large   & 0.08  &  0.02 &  0.20\textsuperscript{*} \\  
     \midrule
  All Base Trans.   & 0.16  & 0.09 & 0.34 \\ 
  All Large Trans.  &  0.09 & -0.01 & 0.23 \\ \midrule 
   BInferSent  & 0.14\textsuperscript{***}  & -0.003\textsuperscript{***} &  0.29\textsuperscript{***} \\ 
   BOW  & 0.02  & 0.03 & 0.04 \\ 
     \bottomrule
    \end{tabular}
    \end{adjustbox}
        \caption{Importance Alignment between model importance scores and oracle importance scores ($\hard$, $\soft$ and $\expert$ metrics) across 5 random seeds on the ANLI dataset. \textbf{$\Delta\mathcal{A}$ was computed over the 60 examples with expert annotations}. `*'s indicate whether $\Delta\mathcal{A}$ is significantly greater than $0$. `***' indicates $p < 0.001$, `**' indicates $p < 0.01 $ and `*' indicates $p < 0.05$. \newline}\label{tab:results_subset}
\end{table}

\paragraph{HANS.}
We measured the extent to which importance alignment scores correlated with accuracy on the HANS diagnostic dataset \cite{mccoy-etal-2019-right}. This dataset measures the extent to which NLI models rely on non-human like heuristics when performing inference, such as inferring that the context entails the hypothesis if all of the words in the hypotheses are in the context: models which do not rely on these non-human like heuristics have higher accuracy on the this dataset. If we assume that na\"ive annotators in general do not expect NLI models to rely on such heuristics, then we might expect models with high importance alignment (i.e., models which make inference decisions in ways annotators expect them to) to also have high accuracy on the HANS dataset.

\section{Results}

\begin{figure*}[ht]
    \centering
    \begin{subfigure}[t]{0.49\textwidth}
        \includegraphics[width=0.95\textwidth]{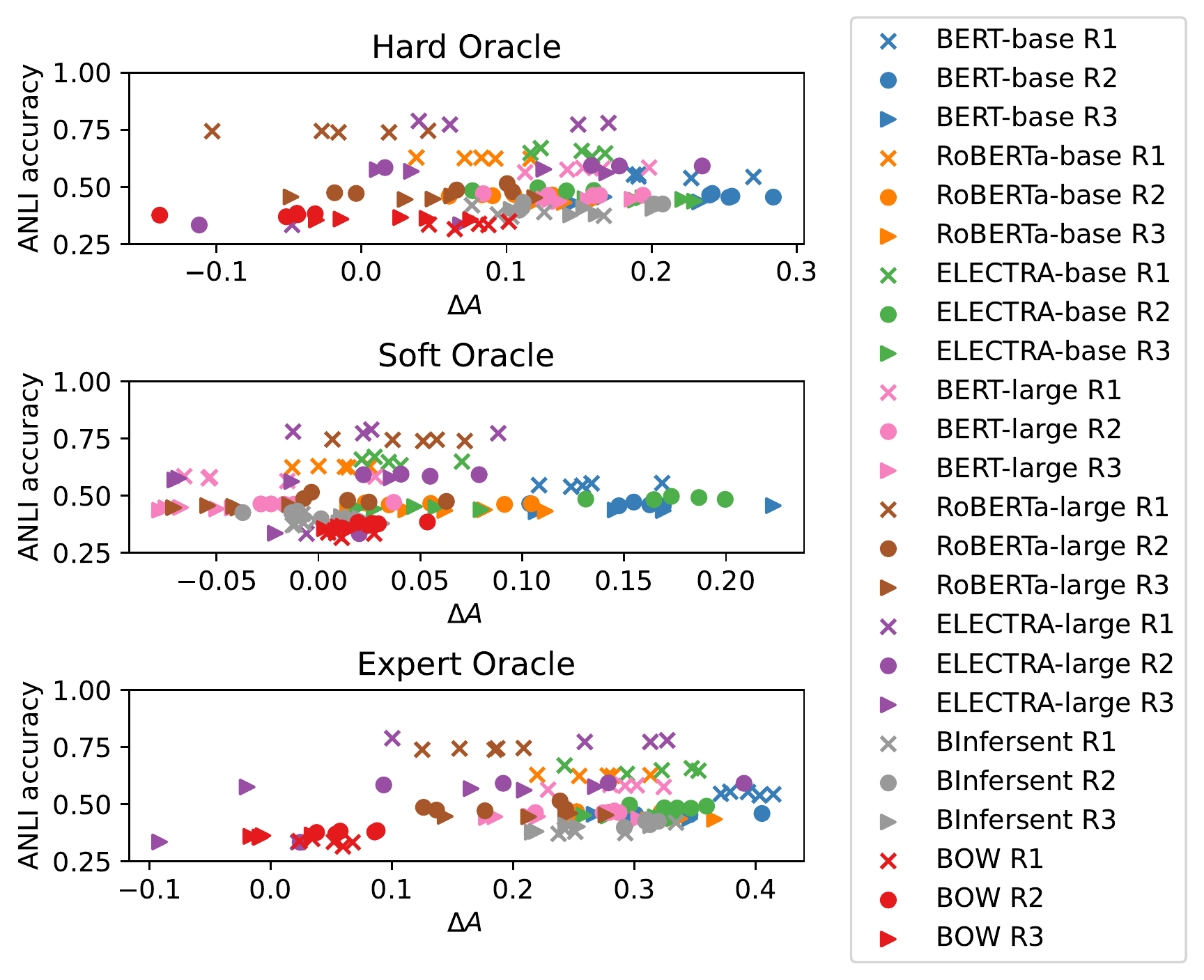}
        \caption{\label{fig:accuracy_and_alignment}Accuracy on ANLI is not correlated with importance alignment ($\mathcal{A}^{H}$ and $\mathcal{A}^{S}$). The cross, circle, and triangle refer to rounds 1, 2, and 3 of ANLI, respectively.}
    \end{subfigure}\hfill
    \centering
    \begin{subfigure}[t]{0.49\textwidth}
      \includegraphics[width=0.95\linewidth]{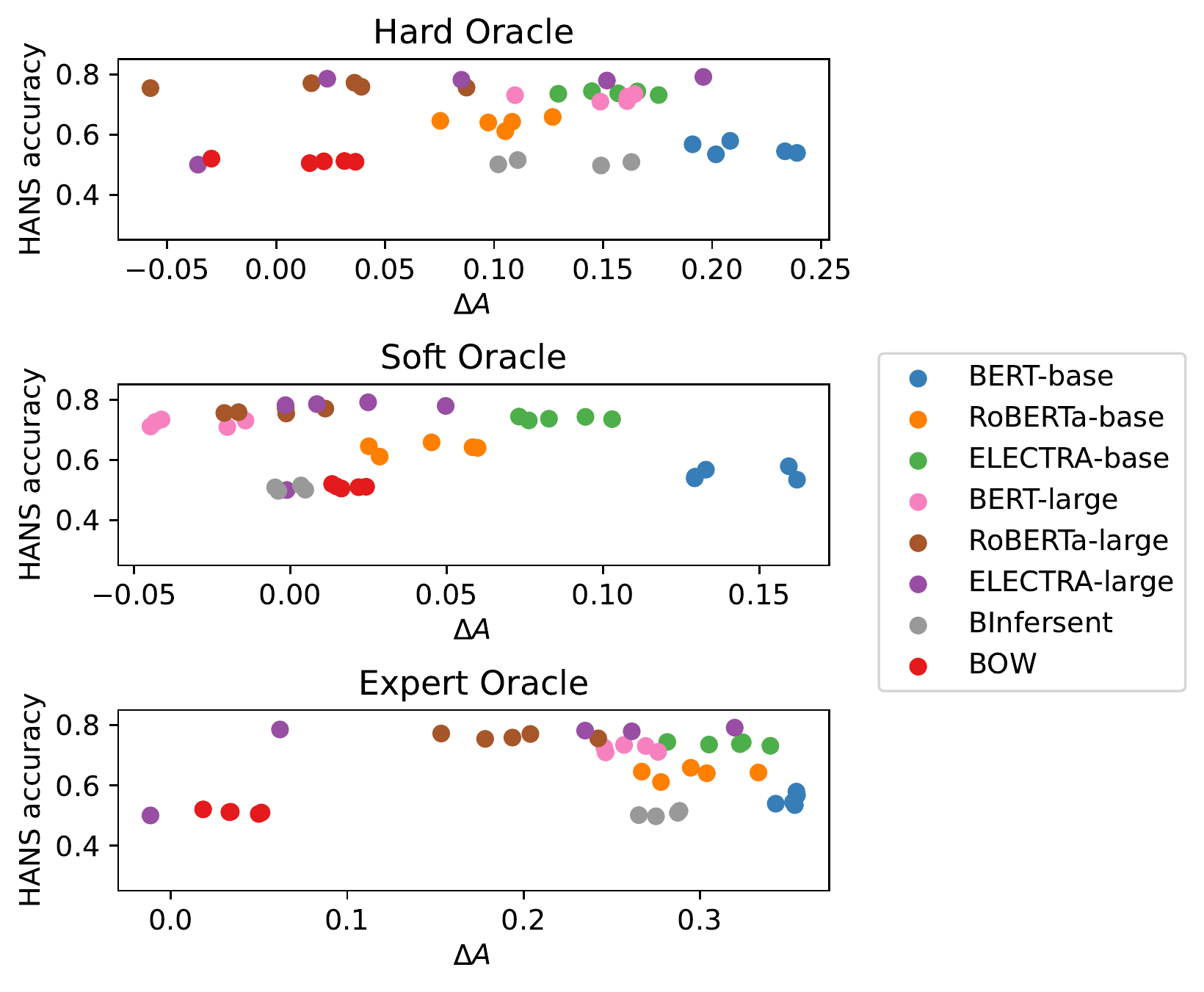}
        \caption{\label{fig:hans_accuracy_and_alignment}Accuracy on HANS is not correlated with importance alignment.  Alignment values are averaged across rounds because HANS is not divided into rounds.}
    \end{subfigure}
\end{figure*}

In \autoref{tab:results}, we report the importance alignment scores for the hard and soft oracles computed over all the examples the models generated an incorrect prediction for, averaged across the five random seeds. In \autoref{tab:results_subset}, we report the importance alignment scores for hard, soft and expert oracles computed over the subset of 60 examples with expert annotations. Since the importance alignment scores for the hard and soft oracles are nearly identical across both tables, we focus our discussion of the results from the subset of examples with expert annotations.
We repeated the reported analyses with hard and soft oracles on all the examples, and found qualitatively similar results (see \autoref{appendix:stats}).

\paragraph{Effect of model size on Importance Alignment.}
Across all three types of Importance Alignment scores ($\hard$, $\soft$ and $\expert$) the base versions of the transformer models had higher importance alignment than their larger counterparts, with BERT-base having the highest importance alignment. To test the statistical significance of this this observation, we fit three linear mixed effect regression models: one for each type of oracle. 

We predicted the pair-wise difference between $C_m(x,y)$ and $C_{m_R}(x,y)$ (see \autoref{eq:cor} and \autoref{eq:cor_rand}) as a function of the following predictors: model size (base vs. large), model type (BERT vs. ELECTRA and BERT vs. RoBERTa) and the interaction between the two. We included model type and its interaction with model size as predictors to measure the effect of model size over and above the differences between specific models. We also included a random intercept and random slope of model size for every example to incorporate the following assumptions: first, the difference $C_m(x,y)$ and $C_{m_R}(x,y)$ can differ for every example; second, the difference between base and large models can also differ for every example. The results described below were significant at a threshold of $p < 0.005$ unless specified otherwise. For further details see \autoref{appendix:stats}. 

The analyses indicated that across all measures importance alignment, $\Delta\mathcal{A}$ was significantly greater for the base models when compared to their larger counterparts. Additionally, $\Delta\mathcal{A}$ was significantly greater in BERT models than in RoBERTa and ELECTRA models (base and large).

Although the results suggest that smaller models have stronger importance alignment than their larger counterparts, our experiments with the BInfersent and BOW models suggest that smaller models do not always have higher importance alignment: the importance alignment for both these models is lower than the alignment for BERT-base model (the smallest of the transformer models when taking into consideration both the number of parameters and pre-training size). 

To test the statistical significance of this numerical result, we fit another set of linear mixed effects model where we predicted the pair-wise difference between $C_m(x,y)$ and $C_{m_R}(x,y)$ as a function of model type (BERT-base vs. BInferSent and BERT-base vs. BOW). We also included a random intercept of item. As expected, the importance alignment for the BERT-base model was significantly greater than that for the BInferSent and BOW models for all the measures.

\paragraph{ANLI accuracy and Importance Alignment.}
Based on the results from \autoref{tab:results} and \autoref{tab:results_subset}, we wondered whether importance alignment might be negatively correlated with NLI accuracy: the base versions of the models which have higher importance alignment have lower accuracy on ANLI compared to their larger counterparts. To test this, we computed a separate value of $\hard$, $\soft$ and $\expert$ for each random seed of the target model and for each round of ANLI. Then, we computed the product moment (i.e., Pearson's) correlation between model accuracy on NLI and $\hard$, $\soft$ and $\expert$. We found a significant correlation between accuracy and $\expert$ ($r=0.42$; $p = 0.008$). However, this correlation was being driven entirely by the BOW models, which had both low accuracy and low importance alignment scores. When repeating the analysis with the BOW models excluded, we found that none of the measures of importance alignment were significantly correlated with NLI accuracy ($\hard$: $r = -0.09$ and $p=0.61$; $\soft$: $r = 0.04$ and $p=0.80$; $\expert$ $r = -0.04$ and $p=0.83$; see \autoref{fig:accuracy_and_alignment}). We repeated the analyses for $\hard$ and $\soft$ for all wrong examples and found that only $\hard$ was significantly correlated with accuracy ($r=0.39$, $p=0.01$; see Appendix~\ref{appendix:corr})

\paragraph{HANS accuracy and Importance Alignment.}
As discussed earlier, we hypothesized that high importance alignment might result in models relying less on non-human-like heuristics, thereby resulting in higher accuracy on the HANS dataset. We found no such correlation, however ($\hard$: $r = 0.04$ and $p=0.82$; $\soft$: $r = -0.21$ and $p=0.21$; $\expert$ $r = 0.23$ and $p=0.17$; see \autoref{fig:hans_accuracy_and_alignment}).\footnote{When we considered the heuristics separately, we found some marginally significant correlations (see Appendix~\ref{appendix:corr})} 

This lack of correlation is likely a result of the mismatch between how human-likeness is defined in HANS and in our importance alignment measures. In HANS, the targeted heuristics are simple (i.e., can be articulated with a rule), and describe general principles of how models ought not behave if they are to be human-like. In contrast, our measures of importance alignment are derived from example level explanations of how na\"ive annotators expected models to behave, and as such are not based on any overarching easy-to-articulate principles. When evaluating whether models make decisions as humans expect them to, jointly considering both these definitions of human-likeness can be useful.

\paragraph{Comparing the Importance Alignment Scores.}
We used hard and soft oracles to automatically measure oracle importance scores. To validate these methods, we computed the Spearman rank correlation between the importance scores derived from these methods and from the manually annotated expert oracle. The hard oracle was moderately correlated with the expert oracle ($r=0.24$, $p<0.0001$), whereas the soft oracle was more weakly correlated ($r=0.14$, $p<0.0001$). Additionally, the hard and soft oracle importance scores were also weakly correlated with each other ($r=0.11$, $p<0.0001$).\footnote{The results are comparable with Pearson's correlation.} Taken together, these results suggest that neither the hard nor the soft oracle measures are perfect proxies for expert human importance scores. This imperfection does not impact the conclusions we draw in this paper, however: the results we discussed held true across all the measures.

\section{Discussion}
In this paper, we argued that it is important to not only evaluate models on how accurate they are on a given task, but also on whether the decisions that the models make align with how humans expect them to make these decisions.
We introduced a measure called Importance Alignment which quantifies the extent to which the parts of the input that non-expert annotators expected to influence models' decisions actually influenced the decisions. To quantify which parts of the input influenced model decisions (\textit{model importance}), we used Integrated Gradients. To quantify annotator expectations (\textit{oracle importance}), we proposed three methods of quantifying annotator generated natural language explanations of model behaviour: two automatic and one that relied on expert input. 

As a case study, we applied this method to measure Importance Alignment in eight NLI models (six transformers, an LSTM and a BOW model), using annotator generated explanations from the ANLI dataset. We found that, across all three measures of importance alignment, the base versions of the transformer models had significantly higher importance alignment than their larger counterparts, with BERT-base having the highest importance alignment. Smaller models do not always result in higher importance alignment, however: the BERT-base model had higher alignment than the LSTM and BOW models. Additionally, importance alignment scores were not correlated with model accuracy on ANLI or the HANS diagnostic dataset in most cases. This suggests that importance alignment and accuracy are complementary methods of evaluating models. 

\paragraph{Future work.} There are at least four directions in which this work can be extended. First, future work can evaluate whether our conclusions about model size in NLI models generalizes to smaller transformer models \citep{turc2019, warstadt-etal-2020-learning} and to models trained on other NLP tasks. 

Second, future work can build on our methods of calculating model and oracle importance. The two methods of automatically computing oracle importance we proposed as a starting point were only moderately correlated with the method that relies of expert input.  
Future work can develop better methods of measuring oracle importance incorporating the strengths of both. 
Future work can also explore other existing ways of calculating model importance (e.g., LIME, SHAP and their variants).

Third, future work can generate more detailed datasets of natural language explanations of model behaviour. For example, in the dataset we used, the explanations the annotators provided were an amalgamation of both why the the label was correct and why the model might have been fooled. Additionally, annotators generated explanations only when the model generated an incorrect output. By disentangling these two types of explanations and collecting explanations for when models generate correct outputs future work can separately study whether models succeed and fail in ways people expect them to. Similarly in the dataset we used, each context-hypothesis-label triplet was associated with only one annotator generated explanation. However, it is possible that there are several explanations of model behaviour that are equally valid. Collecting more explanations per triplet can improve our understanding of how people expect models to succeed and fail. 

Fourth, future work could explore which factors drive higher importance alignment. For example, we observed that transformer models with fewer parameters had higher importance alignment than models with more parameters. Does this finding apply to non-transformer architectures too? Is there a threshold for model parameters, where this inverse relationship between model size and alignment score breaks down? Additionally, can other factors like model architecture, type of training objective or the types of sentences the models were trained on influence importance alignment? Such exploration can not only result in models better aligned with human expectations, but also improve our understanding of what drives importance alignment. 

\section{Conclusion}
We proposed a novel metric, Importance Alignment, to measure the extent to which human-generated explanations of model decisions align with how models actually make these decisions. As a case study, we used this metric to evaluate eight different models trained on NLI and found that the BERT-base model had the highest alignment with human-generated explanations. We also found that our metric was not correlated with model accuracy, suggesting that accuracy and Importance Alignment are complementary ways of evaluating models.

\section*{Acknowledgements} We would like to thank Luke Zettlemoyer, Ana Valeria Gonzalez, the \href{https://dynabench.org/}{Dynabench} team, Roy Schwartz, Oana Maria Camburu, Tom McCoy, and Tal Linzen's Computation and Psycholinguistics lab for their invaluable comments and support.

\bibliography{anthology,references}
\bibliographystyle{acl_natbib}

\newpage

\appendix
\onecolumn

\section{Details about statistical analyses} \label{appendix:stats}
\subsection{Effect of model size}
\noindent Formula for Transformer models
\begin{verbatim}
smf.mixedlm("fisher_cor_diff ~ C(model_size, Treatment(reference='base'))
                               *C(model,Treatment(reference='bert'))",
            data = transformer_dat, groups = transformer_dat["ex_id"],
            re_formula="~model_size")
\end{verbatim}

\vspace{1em}
\noindent Formula for smaller models (BERT-base, InferSent and BOW)
\begin{verbatim}
smf.mixedlm("fisher_cor_diff ~ C(model,Treatment(reference='bert'))",
             data = small_dat, groups = small_dat["ex_id"],
             re_formula="~model_size")
\end{verbatim}

\normalsize
\begin{table}[h]
\centering
\begin{adjustbox}{max width=\linewidth}
    \begin{tabular}{llllllll}
    \toprule
    \bf Fit & \bf Coefficient & \multicolumn{2}{l}{\textbf{All wrong examples}} & \multicolumn{3}{l}{\textbf{Expert annotated subset}} \\
      &     &  $\Delta\hard$ &  $\Delta\soft$ &  $\Delta\hard$ &  $\Delta\soft$ & $\Delta\expert$  \\ 
           
   \midrule
   
   Transformer models & Large (vs. Base) & -0.32\threestar & -0.142\threestar &      -0.057\twostar & -0.173\threestar & -0.115\threestar  \\
                      & ELECTRA (vs. BERT) & -0.044\threestar & -0.057\threestar & -0.048 \twostar & -0.067\threestar & -0.054\onestar  \\
                      & RoBERTa (vs. BERT) & -0.099\threestar & -0.101\threestar & -0.106\threestar & -0.105\threestar & -0.083\threestar  \\
                      & ELECTRA : Large & -0.070\threestar & 0.085\threestar & -0.014 & 0.109\threestar & -0.051   \\
                      & RoBERTa : Large & -0.044\threestar & 0.127\threestar & -0.016 & 0.134\threestar & 0.012  \\
   
   \midrule
   
   Smaller models & BOW (vs. BERT-base) & -0.203\threestar & -0.104\threestar  & -0.184\threestar & -0.125\threestar & -0.184\threestar \\
                  & BInferSent (vs. BERT-base) & -0.094\threestar & -0.120\threestar  & -0.080\threestar & -0.154\threestar  & -0.080\threestar\\

     \bottomrule
    \end{tabular}
    \end{adjustbox}
        \caption{\threestar, \twostar and \onestar indicate p < 0.0001, 0.001 and 0.01 respectively and `:' indicates an interaction effect. A separate mixed effects regression model was fit for each column. Negative coefficients for the main effects indicate that the baseline value was greater than the comparison (e.g., Base was greater than Large).}\label{tab:stats_model_size}
\end{table}

\vspace{-1em}

\newpage
\subsection{Correlation with accuracy}\label{appendix:corr}
\normalsize
\begin{table}[h]
\centering
\begin{adjustbox}{max width=\linewidth}
    \begin{tabular}{lllllll}
    \toprule
    \bf Evaluation dataset & \multicolumn{2}{l}{\textbf{All wrong examples}} & \multicolumn{3}{l}{\textbf{Expert annotated subset}} \\
           &  $\Delta\hard$ &  $\Delta\soft$ &  $\Delta\hard$ &  $\Delta\soft$ & $\Delta\expert$  \\ 
           
   \midrule
   
   ANLI Dev & 0.39\onestar & 0.10  & 0.23 & 0.10 & 0.42\twostar (-0.04) \\
   \midrule
   HANS (all)  & 0.18 & -0.19 & 0.04 & -0.21 & 0.22 \\
   HANS (constituent)  & 0.30$^{+}$ & -0.11 & -0.19 & -0.26 & -0.18 \\
   HANS (lexical overlap)  & 0.20 & -0.17 & -0.25 & -0.27 & -0.19 \\
   HANS (subsequence)  & -0.006 & -0.31$^{+}$ & -0.29$^{+}$ & -0.39\onestar & -0.27 \\

     \bottomrule
    \end{tabular}
    \end{adjustbox}
        \caption{`$^{+}$', `\onestar' and `\twostar' indicate p < 0.1, 0.05 and 0.01 respectively. Values in parentheses indicate correlation without the BOW models in cases where they were outliers (see \autoref{fig:accuracy_and_alignment})}\label{tab:stats_accuracy}
\end{table}

\begin{figure}[h]
    \centering
    \begin{subfigure}[t]{0.49\textwidth}
        \includegraphics[width=0.9\textwidth]{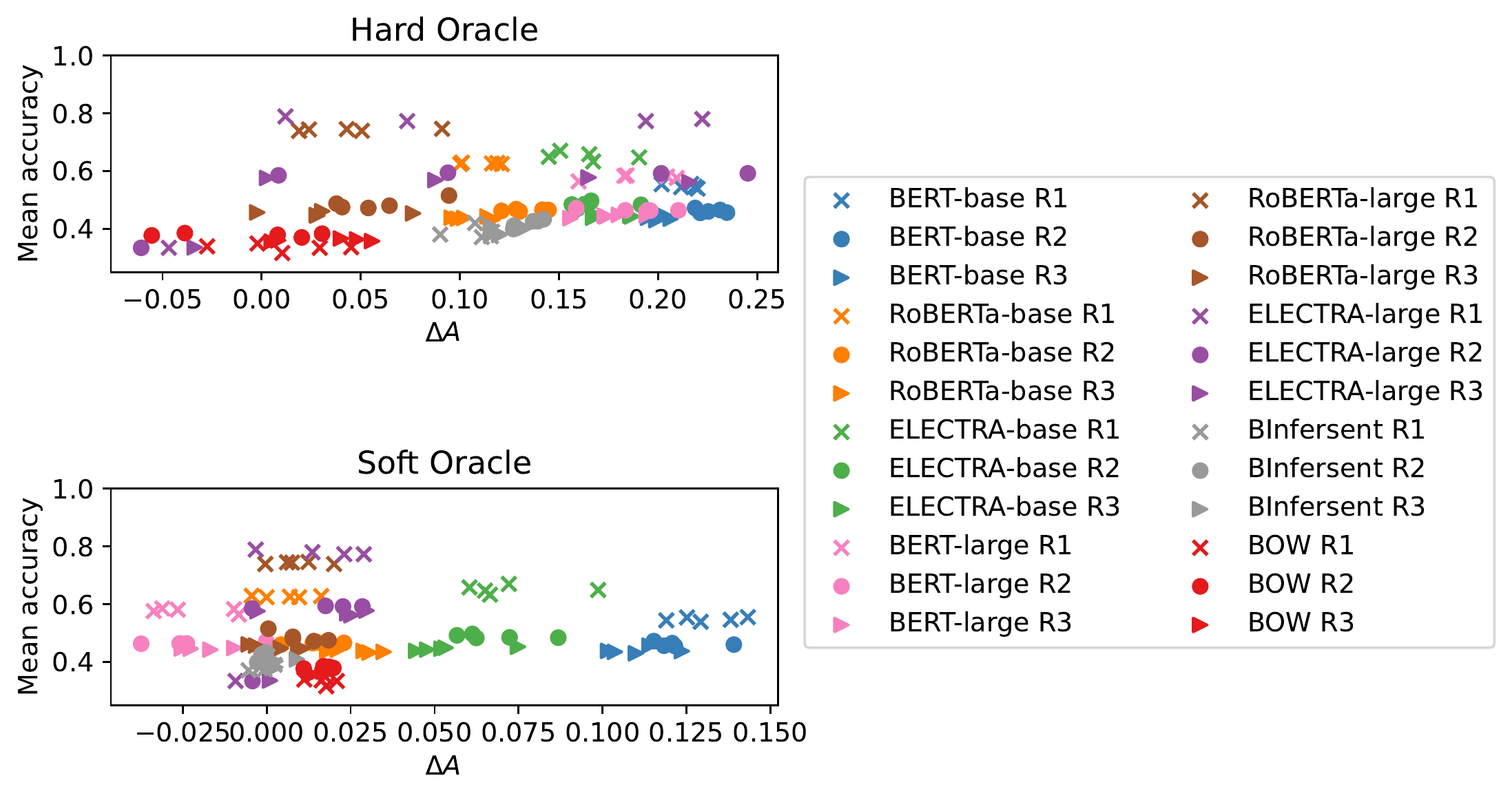}
        \caption{\label{fig:accuracy_and_alignment_all}Correlationg accuracy on ANLI with importance alignment ($\mathcal{A}^{H}$ and $\mathcal{A}^{S}$) for all wrong examples. The cross, circle, and triangle refer to rounds 1, 2, and 3 of ANLI, respectively.}
    \end{subfigure}\hfill
    \begin{subfigure}[t]{0.49\textwidth}
      \includegraphics[width=0.7\linewidth]{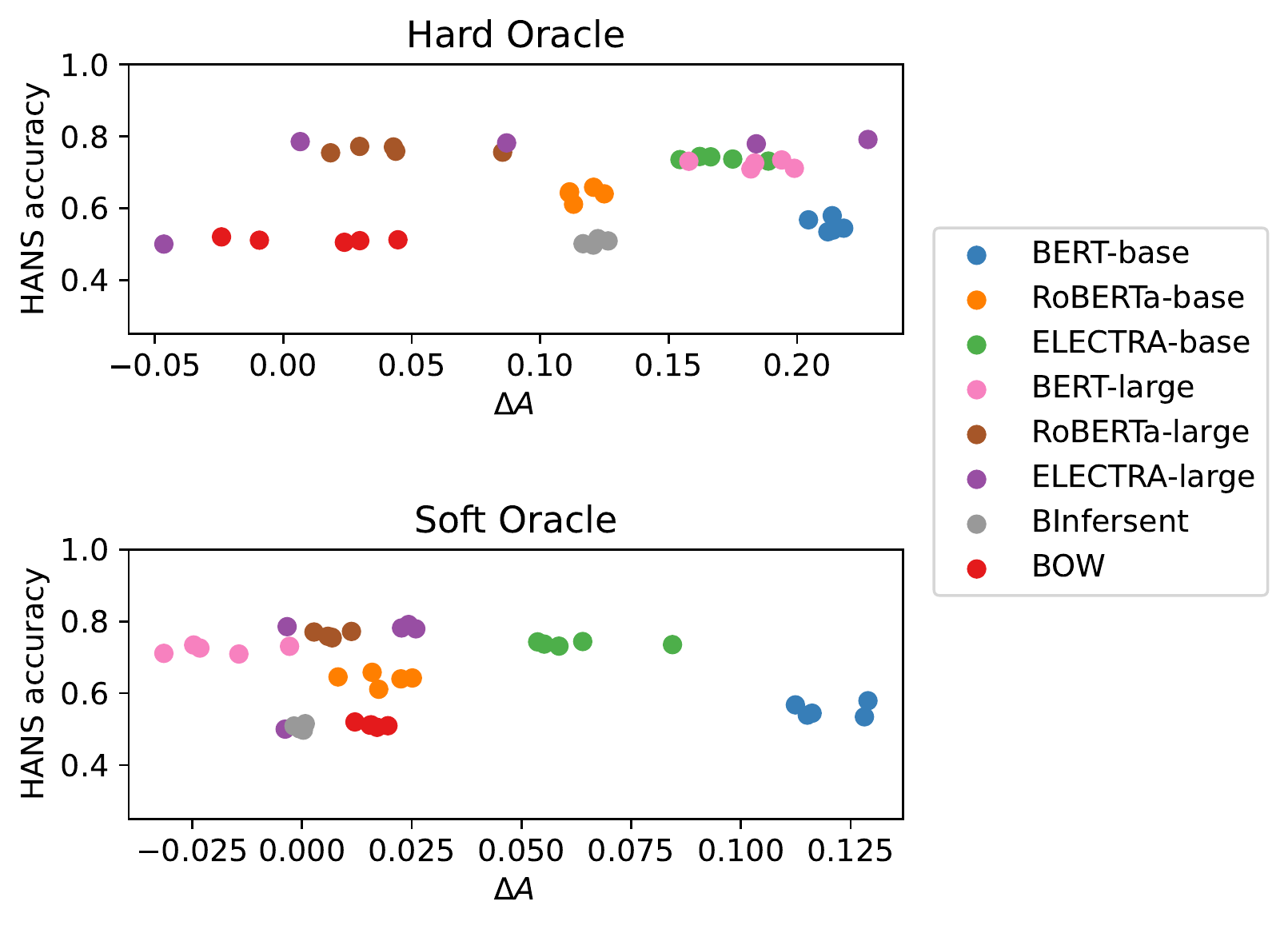}
        \caption{\label{fig:hans_accuracy_and_alignment_all}Correlating accuracy on HANS with importance alignment for all wrong examples.  Alignment values are averaged across rounds because HANS is not divided into rounds.}
    \end{subfigure}
    \caption{}
\end{figure}

\begin{figure}[h]
    \centering
    \begin{subfigure}[t]{0.49\textwidth}
        \includegraphics[width=\textwidth]{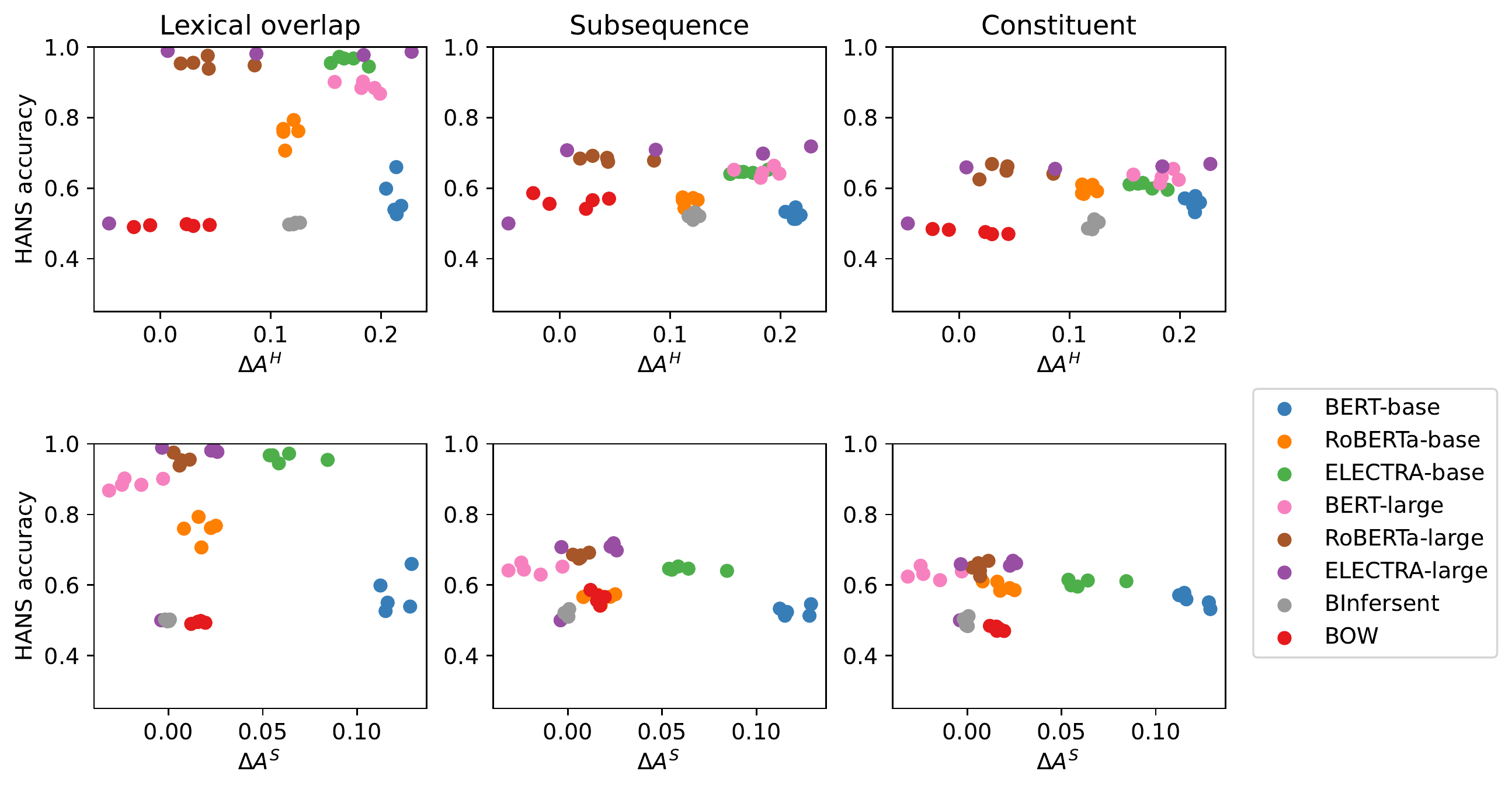}
        \caption{\label{fig:hans_accuracy_and_alignment_all_heuristics}Correlating accuracy on ANLI with importance alignment ($\mathcal{A}^{H}$ and $\mathcal{A}^{S}$) for all wrong examples. The cross, circle, and triangle refer to rounds 1, 2, and 3 of ANLI, respectively.}
    \end{subfigure}\hfill
    \begin{subfigure}[t]{0.49\textwidth}
      \includegraphics[width=\linewidth]{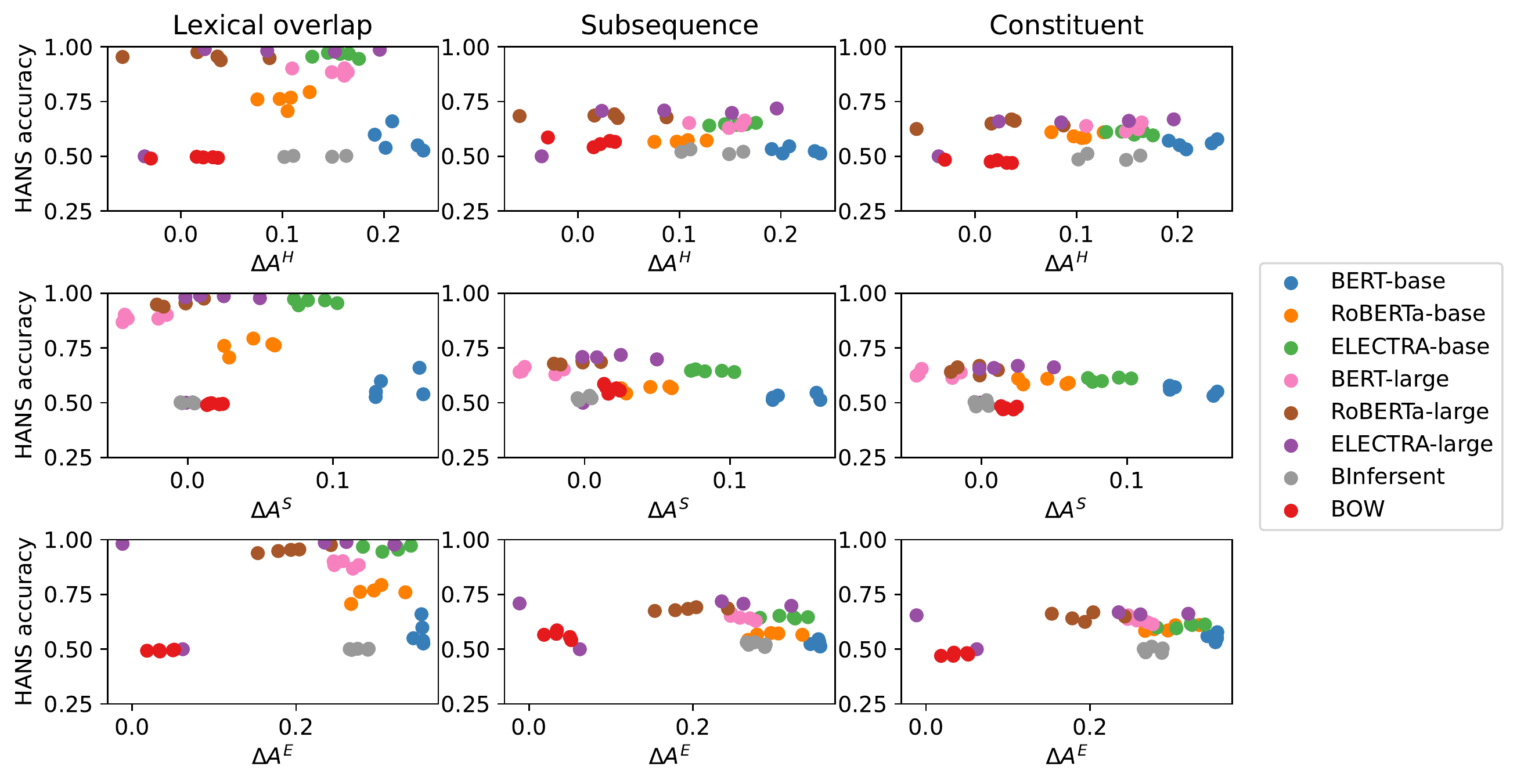}
        \caption{\label{fig:hans_accuracy_and_alignment_subset_heuristics}Correlating accuracy on HANS with importance alignment for all wrong examples.  Alignment values are averaged across rounds because HANS is not divided into rounds.}
    \end{subfigure}
    \caption{}
\end{figure}

\end{document}